\documentclass[conference]{IEEEtran}
\usepackage{cite}
\usepackage{amsmath,amssymb,amsfonts}
\usepackage{algorithmic}
\usepackage{graphicx}
\usepackage{textcomp}
\usepackage{xcolor}
\def\BibTeX{{\rm B\kern-.05em{\sc i\kern-.025em b}\kern-.08em
    T\kern-.1667em\lower.7ex\hbox{E}\kern-.125emX}}
\usepackage{csquotes}
\usepackage{tabu}
\usepackage{tabularx}
\usepackage{threeparttable}
\usepackage{balance}
\begin{document}

\title{Evaluating the validity of a German translation of an uncanniness questionnaire
}

\author{\IEEEauthorblockN{1\textsuperscript{st} Sarah Wingert}
\IEEEauthorblockA{\textit{Computer Science and Media department} \\
\textit{Stuttgart Media University}\\
Stuttgart, Germany \\
0000-0001-5698-3809}
\and
\IEEEauthorblockN{2\textsuperscript{nd} Christian Becker-Asano}
\IEEEauthorblockA{\textit{Computer Science and Media department} \\
\textit{Stuttgart Media University}\\
Stuttgart, Germany \\
0000-0002-9946-0458}
}

\maketitle

\begin{abstract}
When researching on the acceptance of robots in Human-Robot-Interaction the Uncanny Valley needs to be considered. Reusable and standardized measures for it are essential. In this paper one such questionnaire got translated into German. The translated indices got evaluated (n=140) for reliability with Cronbach's alpha. Additionally the items were tested with an exploratory and a confirmatory factor analysis for problematic correlations. The results yield a good reliability for the translated indices and showed some items that need to be further checked.
\end{abstract}


\section{Introduction}
When developing robots that are intended for Human-Robot-Interaction (HRI) it is important that they are being accepted by the target group. It has been found that anthropomorphic features support the social acceptance of the robot given the design is appropriate for its task~\cite{psychologicalViewOnRobots}. However, as Mori suggested, the anthropomorphic design might lead to a feeling of repulsion if too close to an actual human. This effect is being called the \enquote{Uncanny Valley}~\cite{morisUncannyValley}. 

Since the social acceptance of a robot cannot directly be determined by objective measures
, it is necessary to have appropriate tools to measure the acceptance or contrary the eeriness of a robot within the target group~\cite{GODSpeedQuestionnaire}. For research regarding the Uncanny Valley it is also useful to have a measure for human-likeness. Because most engineers are not trained to create valid questionnaires nor have the capacities to validate created questionnaires properly, standardized measures that can be reused between research are of great value~\cite{GODSpeedQuestionnaire}. Two such questionnaires 
are the GODSpeed-questionnaire~\cite{GODSpeedQuestionnaire} and the questionnaire by MacDorman and Ho ~\cite{questionnaireHo} that specifically addresses research on the Uncanny Valley.

The GODSpeed-questionnaire contains five parts measuring the concepts of anthropomorphism, animacy, likeability, perceived intelligence and perceived safety. Each concept contains three to six items presented in the form of semantic differentials~\cite{GODSpeedQuestionnaire}. At the time of writing the GODSpeed-questionnaire is available in 18 different languages~\cite{GODSpeedWebsite}. This enables the use of the questionnaire in different cultures and language areas.
However, when researching the Uncanny Valley, the lack of an index measuring eeriness is problematic, as it is inherently different from reverse likeability and essential for identifying an Uncanny Valley. Additionally the indices contained in the GODSpeed-questionnaire are significantly correlated with each other, which leads to a highly skewed diagram, if two indices of the set are used as x- and y-axis of one chart~\cite{questionnaireHo}.

Since the correlation between the indices in the GODSpeed-questionnaire got traced back to the correlation of each with interpersonal warmth, MacDorman and Ho created indices that are decorrelated from interpersonal warmth and from each other, to enable the measurement of less skewed data when researching the Uncanny Valley~\cite{questionnaireHo}. This questionnaire got revised again with a broader range of sample data to avoid clustering in the responses~\cite{questionnaireHoRevised}.


\section{Translation of the questionnaire to German}
\label{sec-translation}

\begin{table}[h]
	\centering
	\tabulinesep=3mm
	\begin{threeparttable}
	\caption{The translations for humanlikeness}\label{transHuman}
	\begin{tabularx}{0.49\textwidth}{|X| X | X| X |}
	\hline
	\textbf{English} & \multicolumn{2}{|c|}{\textbf{German}} & \textbf{English} \\
	\hline
	inanimate & unbeseelt & lebendig & living \\
	\hline
	synthetic & synthetisch & echt & real \\
	\hline
	mechanical \newline movement & mechanische \newline Bewegungen & biologische \newline Bewegungen & biological \newline movement \\
	\hline
	human-made & von Menschen \newline gemacht & wie ein Mensch & human-like \\
	\hline
	without definite lifespan & mit unbegrenzter Lebenszeit* & sterblich* & mortal \\
	\hline
	\hline
	\textit{artificial} & \textit{künstlich} & \textit{natürlich} & \textit{natural} \\
	\hline
	\end{tabularx}
	\begin{tablenotes}
        \item[\textit{italic}] plausibility question
        \item[*] outstanding in analysis, further checks/adaptations needed
    \end{tablenotes}	
	\end{threeparttable}
\end{table}

In our work we researched on the effect of three different faces on the perception of a mobile robot~\cite{baThesis}. The research specifically focused on the question whether a \enquote{Perceptual Mismatch} between robot and face would lead to a high eeriness as proposed in the literature~\cite{CreepyCatsAndStrangeHighHouses}, when measures were taken to ensure, that 
the robot and the face are processed individually. 
To avoid distortion of the results it is important that the axes for humanlikeness and eeriness are decorrelated. Therefore the revised questionnaire of MacDorman and Ho was chosen.
However, as the work was performed in Germany, a German version of the questionnaire was required. To the best of our knowledge, at the time this translation was done, there was no other German translation available. Hence we carried out the translation ourselves, cf.~Table~\ref{transHuman} to \ref{transAttract}. However, we recently came across another German translation~\cite{germanTranslation}.

Generally the back translation method is advised to ensure a proper translation of a questionnaire into another language~\cite{GODSpeedQuestionnaire}. In this method one bilingual person translates the original questionnaire into the target language, while a second bilingual person translates the target language back into the source language without knowing the original version. If the result of the back translation is equivalent to the original source, it can be assumed that the translation is equivalent as well~\cite{backTranslation}.

\begin{table}[h]
	\centering
	\tabulinesep=3mm
	\begin{threeparttable}
	\caption{The translations for eeriness}\label{transEeriness}
	\begin{tabularx}{0.49\textwidth}{|X| X | X| X |}
	\hline
	\textbf{English} & \multicolumn{2}{|c|}{\textbf{German}} & \textbf{English} \\
	\hline
	dull & reizlos & irre & freaky \\
	\hline
	predictable & vorhersehbar & unheimlich & eerie \\
	\hline
	plain & eindeutig & komisch & weird \\
	\hline
	ordinary & gewöhnlich & ungewöhnlich & supernatural \\
	\hline
	boring & langweilig & schockierend & shocking \\
	\hline
	uninspiring & wenig anregend & schaurig & spine-tingling \\	
	\hline
	predictable & vorhersehbar & spannend & thrilling \\
	\hline
	bland & fade & unheimlich & uncanny \\
	\hline
	unemotional & löst keine \newline Emotionen aus & haarsträubend & hair-raising \\
	\hline
	\hline
	\textit{reassuring} & \textit{beruhigend*} & \textit{unheimlich*} & \textit{eerie} \\
	\hline
	\end{tabularx}
	\begin{tablenotes}
        \item[\textit{italic}] plausibility question
        \item[*] outstanding in analysis, further checks/adaptations needed
    \end{tablenotes}	
	\end{threeparttable}
\end{table}

In our translation the back translation approach was successful for the humanlikeness index, cf.~Table~\ref{transHuman}. Two independent back translators each translated 80\% of the words back identically as the source, leading to a total coverage of 90\% identical translations. The wording \enquote{without definite lifetime} didn't get translated back exactly, but the back translation of \enquote{unlimited lifetime} was interpreted as equivalent. Also the plausibility question was translated back identically by one back translator.

In the index of eeriness a few adaptations were made, cf.~Table~\ref{transEeriness}. As the English questionnaire used some words that in German would describe feelings evoked by the robot mixed with words directly describing the robots perception (e.g. \enquote{dull} vs. \enquote{freaky}) the translations were adapted to equivalent words describing the robots perception. For example the German translation for \enquote{dull} is a direct translation of its synonym \enquote{plain} instead of the word itself. Despite the adaptations particular attention was paid to the fact that the connotations with positive (and negative) affect within one semantic differential didn't change. As English has a variety of words describing emotional states and the translations were singular words without a broader context, the back translation process was not viable for emotional states. Therefore, the index was checked for equivalent meanings by only one bilingual native speaker proofreading the source and target translation. After careful consideration the translation of \enquote{weird} with \enquote{komisch} was identified as a double meaning, because the German word can also mean \enquote{funny}. However, there were no noticeable problems with it in the analysis. An adaption to the alternative word \enquote{seltsam} might still be advisable.

\begin{table}[h]
	\centering
	\tabulinesep=1mm
	\begin{threeparttable}
	\caption{The translations for attractiveness}\label{transAttract}
	\begin{tabularx}{0.5\textwidth}{|X| X | X| X |}
	\hline
	\textbf{English} & \multicolumn{2}{|c|}{\textbf{German}} & \textbf{English} \\
	\hline
	ugly & hässlich & schön & beautiful \\
	\hline
	crude & grob & stylisch & stylish \\
	\hline
	repulsive & abstoßend & angenehm & agreeable \\
	\hline
	messy & unschön & geschmeidig & sleek \\
	\hline
	\hline
	\textit{unattractive} & \textit{unattraktiv} & \textit{attraktiv} & \textit{attractive} \\
	\hline
	\end{tabularx}
	\begin{tablenotes}
        \item[\textit{italic}] plausibility question
    \end{tablenotes}	
	\end{threeparttable}
\end{table}

The index of attractiveness got translated, but not checked, because it wasn't part of the statistical analysis of our work. The translations are reported in Table~\ref{transAttract}, but using the back translation method for confirmation is advised.

\begin{figure}[h]
	\begin{center}
		\includegraphics[width=0.5\textwidth]{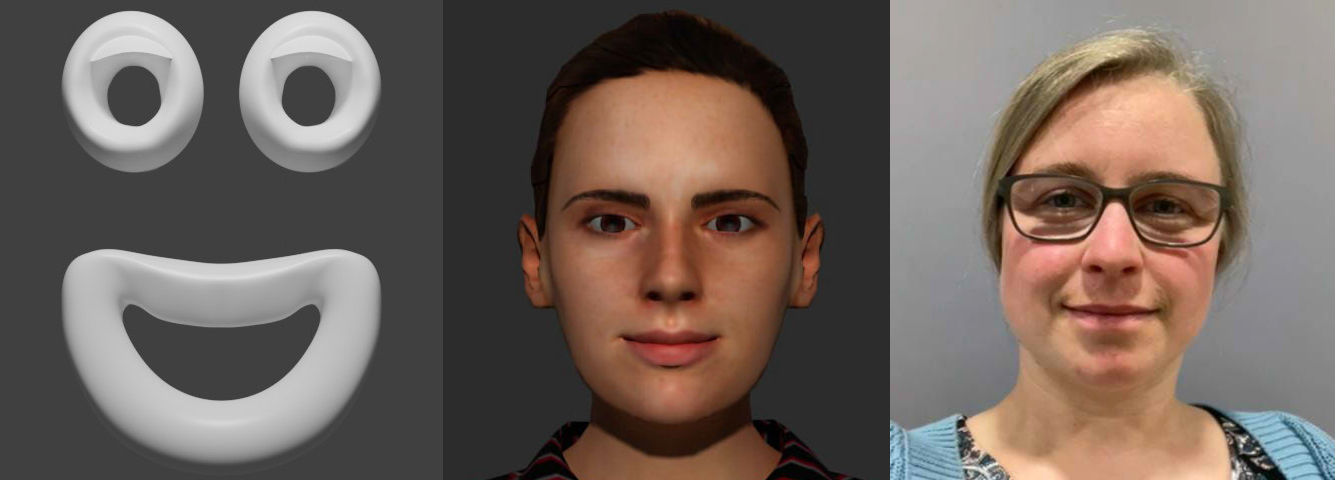}
		\caption{The faces in the neutral position: from left to right, face 1, face 2, and face 3}
		\label{fig2}
	\end{center}
\end{figure}

\section{Validation and reliability testing}
As mentioned in Section \ref{sec-translation}, our German version of the questionnaire was used to evaluate, how the humanlikeness and eeriness of a robot's physical appearance are changing with different versions of facial display of emotions. 

\begin{figure*}[h!]
\centering
    \includegraphics[width=0.8\textwidth]{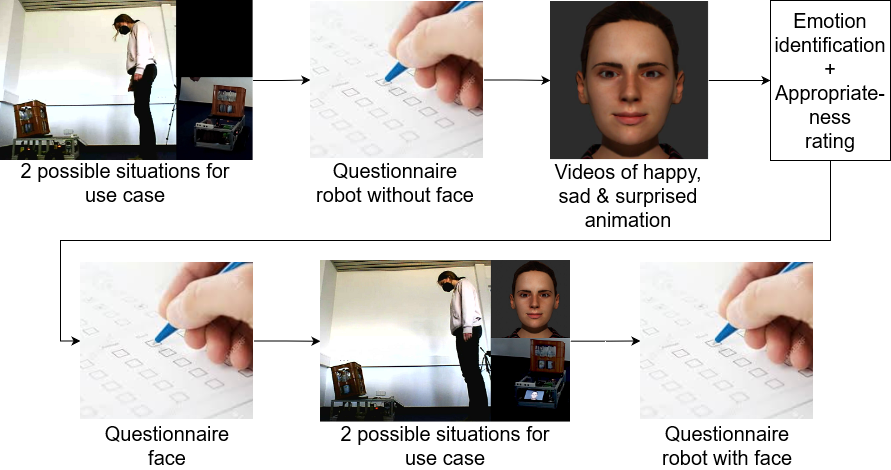}
    \caption{The experimental procedure}
    \label{figProcedure}
\end{figure*}

\begin{figure}[h!]
		\includegraphics[width=0.24\textwidth]{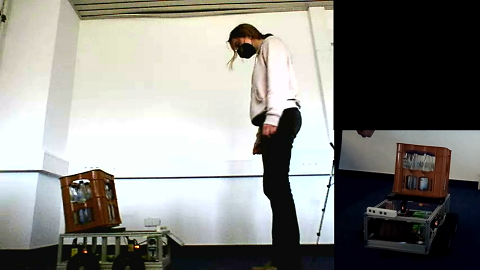}\hfill
		\includegraphics[width=0.24\textwidth]{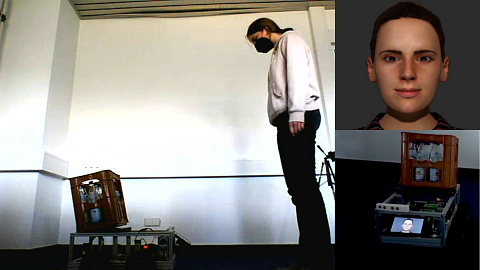}
		\caption{Two exemplary frames of the videos watched by the participants: left, robot without a face, and right, robot with face 2}
  \label{fig1}
\end{figure}
The robots use case was the transportation and sale of food and water bottles. A human confronting the robot could either buy food or not. To support anthropomorphisation and emotional bonding to the robot, emotional reactions were implemented. When a human approached, the robot showed surprise. When the human bought something, the robot showed happiness, and if nothing was bought, it showed sadness. These emotions were modelled with three faces with different levels of humanlikeness, depending on the test group, cf.~Figure~\ref{fig2}.

\subsection{Data collection procedure}
The survey was conducted online with a series of videos being presented to each participant, see Figure~\ref{figProcedure} for an overview. Following a between-subjects experimental design the participants were split into three groups. Each group watched only one version of the animated face. The participants were recruited at three technical universities and one IT-company in Germany. Sixty-seven responses were returned, of which six were excluded, because they didn't answer the control question correctly, and one person was excluded for watching none of the videos. This resulted in 60 complete surveys for our analysis. After exclusion of mentioned participants, the first group contained 21 participants (face 1 condition), the second group 19 (face 2 condition), and the third group 20 (face 3 condition). 43.3\% of the participants were female and 56.7\% male. Overall the participants were rather young (M=28.0 years; SD=8.51 years). 46.7\% had at least one university degree and 91.7\% had at least an admission to higher education.

The sequence of the steps of the online survey is presented in Fig.~\ref{figProcedure}. Each participant first watched the robot alone (with its display turned off, cf.~Fig.~\ref{fig1}, left). One video showed the system when something was bought, the other video showed it when nothing was bought. This covered the whole use case of the robot. Afterwards the participants were asked to rate the robot with the questionnaire. In the next step each test group watched videos of the face corresponding to their group displaying the three emotions mentioned above. They were asked to identify the emotions and rate their appropriateness. Afterwards an image of the face in neutral position was given and the participants were asked to rate the overall perception of the face on the questionnaire. Finally each group watched the robot in combination with the respective face shown on the screen in the same situations as in the beginning, cf.~Fig.~\ref{fig1}, right. After that they rated this combination of robot and face filling in the German questionnaire once more.\footnote{The results of the study itself will be reported at a later time.}

The combination of the data of the three groups resulted in a total of 180 ratings, of which 60 were ratings of the robot alone, 21 of face 1 alone, 19 of face 2 alone, 20 of face 3 alone, and equivalent amounts of ratings for the combinations of each face with the robot. For validation of the questionnaire's translation, 20 of the 60 ratings of the robot were chosen at random so that each condition provided a similar amount of data. Thus, in total 140 data sets are included in the reliability testing. 

\begin{table*}[h!]
	\centering
	\tabulinesep=3mm
	\begin{threeparttable}
	\caption{Cronbach’s Alpha for the question items}\label{cronbachsAlpha}
	\begin{tabularx}{0.8\textwidth}{|>{\hsize=0.2\hsize}X| >{\hsize=1.5\hsize}X|*{3}{>{\hsize=0.7\hsize}X|}}
	\hline
		No. & Item & \multicolumn{3}{c|}{Cronbach’s~$\alpha$ if item dropped} \\
		\hline
		& & Human- \newline likeness & Eeriness & Attractiveness\\
		\hline
	 H1 & unbeseelt – lebendig & 0.887 & & \\
	 H2 & synthetisch – echt & 0.871 & & \\
	 H3 & mechanische Bewegungen –  biologische Bewegungen & 0.880 & & \\
	 H4 & von Menschen gemacht – wie ein Mensch & 0.870 & & \\
	 H5 & mit unbegrenzter Lebenszeit – sterblich* & 0.916 & & \\
	 H6 & \textit{künstlich – natürlich} & \textit{0.869} & &\\
	\hline
	E1 & reizlos – irre & & 0.798 & \\
	E2 &vorhersehbar - unheimlich & & 0.778 & \\
	E3 & eindeutig - komisch & & 0.795 & \\
	E4 & gewöhnlich – ungewöhnlich & & 0.802 & \\
	E5 &langweilig – schockierend & & 0.794 & \\
	E6 & wenig anregend – schaurig & & 0.773 & \\
	E7 & vorhersehbar – spannend & & 0.805 & \\
	E8 & fade – unheimlich & & 0.780 & \\
	E9 & löst keine Emotionen aus – haarsträubend & & 0.781 & \\
	E10 & \textit{beruhigend – unheimlich} & & \textit{0.792} & \\
	\hline
	A1 & hässlich – schön & & & 0.845 \\
	A2 & grob – stylisch & & & 0.862 \\
	A3 & abstoßend – angenehm & & & 0.870 \\
	A4 & unschön – geschmeidig & & & 0.847 \\
	A5 & \textit{unattraktiv – attraktiv} & & & \textit{0.844} \\
	\hline
	\hline
	& Cronbach’s $\alpha$ for factor & 0.901 & 0.807 & 0.880 \\
	\hline
	\end{tabularx}
	\begin{tablenotes}
        \item[\textit{italic}] plausibility question
        \item[*] outstanding in analysis, further checks/adaptations needed
        \cite{exploratoryFactorAnalysis}
    \end{tablenotes}	
	\end{threeparttable}
\end{table*}

\subsection{Reliability analysis}
With the resulting data set Cronbach's Alpha~\cite{exploratoryFactorAnalysis} was determined for each of the indices and under exclusion of single question items to check for reliability. In addition an exploratory factor analysis and a confirmatory factor analysis were performed to identify any severe issues appearing in the translation. All analyses were performed using Jamovi~\cite{jamovi}.

Cronbach's alpha indicated a good reliability for all the indices. Humanlikeness had a value of 0.901, eeriness a value of 0.807 and attractiveness a value of 0.880 (see Table~\ref{cronbachsAlpha}). Cronbach's alpha of the humanlikeness index under exclusion of the German translation of the pair \enquote{without definite lifetime}-\enquote{mortal} was slightly higher (\(0.916 > 0.901\)). An analysis revealed that this pair showed a negative correlation with the other items in the data sets of the robot. Due to its design the robot seemed to appear fragile to the participants, which they might have interpreted as mortal. This needs to be considered in further use of the German version of the survey.

\subsection{Exploratory analysis}
The exploratory factor analysis showed that some items correlated with the factor assigned to the attractiveness items (see Table~\ref{efa}, items marked with an asterisk). The item \enquote{gewöhnlich} - \enquote{ungewöhnlich} had a strong bias and also loaded on the humanlikeness factor. As this issue was already mentioned in the original questionnaire~\cite{questionnaireHoRevised}, it is not being considered an issue of the German translation. The uniqueness of the items in the humanlikeness index are relatively low. This indicates that they are strongly correlated with each other. One exception is the item \enquote{mit unbegrenzter Lebenszeit}-\enquote{sterblich} which can be explained by its negative correlation with the other items in the ratings of the robot and supports the exclusion of this item. Most factors of eeriness have a relatively high uniqueness as can be derived from their low correlations with each other. However, their common variance is still a little less than 0.5. The low correlations can be explained by the two sub-factors contained within the set of eeriness items. Since the items only load on one of the two sub-factors, but were summarized in this exploratory factor analysis, high values of uniqueness result. The uniqueness values of the attractiveness index are slightly higher than the items of humanlikeness. Still they are all below 0.5, which supports that the items are well correlated with each other.

\begin{table*}[h!]
	\centering
	\tabulinesep=3mm
	\begin{threeparttable}
	\caption{Exploratory factor analysis for the question items}\label{efa}
	\begin{tabularx}{0.8\textwidth}{|>{\hsize=0.2\hsize}X| >{\hsize=1.6\hsize}X|*{3}{>{\hsize=0.4\hsize}X}| >{\hsize=0.8\hsize}X|}
	\hline
	No.	& & \multicolumn{3}{c|}{Factor} & \\
		\cline{3-5}
		& & 1 & 2 & 3 & Uniqueness \\
		\hline
	 H1 & unbeseelt – lebendig* & 0.534 & & 0.384 & 0.342 \\
	 H2 & synthetisch – echt & & & 0.758 & 0.280 \\
	 H3 & mechanische Bewegungen –  \newline biologische Bewegungen* & 0.376 & & 0.554 & 0.325 \\
	 H4 & von Menschen gemacht – wie ein Mensch & & & 0.765 & 0.256 \\
	 H5 & mit unbegrenzter Lebenszeit – sterblich & & & 0.683 & 0.645 \\
	 H6 & \textit{künstlich – natürlich} & & & \textit{0.783} & 0.234 \\
	\hline
	 E1 & reizlos – irre & & 0.484 & & 0.685 \\
	E2 & vorhersehbar - unheimlich & & 0.650 & & 0.579 \\
	E3 & eindeutig - komisch & & 0.512 & & 0.612 \\
	E4 & gewöhnlich – ungewöhnlich* & 0.414 & 0.358 & -0.552 & 0.602 \\
	E5 & langweilig – schockierend & & 0.500 & & 0.691 \\
	E6 & wenig anregend – schaurig & & 0.734 & & 0.467 \\
	E7 & vorhersehbar – spannend* & 0.487 & 0.361 & & 0.682 \\
	E8 & fade – unheimlich & & 0.669 & & 0.549 \\
	E9 & löst keine Emotionen aus – haarsträubend & & 0.684 & & 0.539 \\
	E10 & \textit{beruhigend – unheimlich*} & \textit{-0.488} & \textit{0.551} & & \textit{0.344} \\
	\hline
	A1 & hässlich – schön & 0.686 & & & 0.429 \\
	A2 & grob – stylisch & 0.602 & & & 0.450 \\
	A3 & abstoßend – angenehm & 0.623 & & & 0.414 \\
	A4 & unschön – geschmeidig & 0.726 & & & 0.364 \\
	A5 & \textit{unattraktiv – attraktiv} &  \textit{0.771} & & & 0.288 \\
	\hline
	\end{tabularx}
	\begin{tablenotes}
        \item[\textit{italic}] plausibility question
        \item[*] correlation with factor of attractiveness
        \newline Created with Principal Axis Extraction and 'oblimin' rotation
        \newline Loadings $<0.3$ not mentioned
       	\cite{exploratoryFactorAnalysis}
    \end{tablenotes}	
	\end{threeparttable}
\end{table*}

\subsection{Confirmatory analysis}
The confirmatory factor analysis confirmed that all items loaded significantly on their factor ($p<0.01$), cf.~Table~\ref{cfa}. 

\begin{table*}[h!]
	\centering
	\tabulinesep=3mm
	\begin{threeparttable}
	\caption{Confirmatory factor analysis for the question items}\label{cfa}
	\begin{tabularx}{0.8\textwidth}{|>{\hsize=0.2\hsize}X| >{\hsize=1.5\hsize}X|*{3}{>{\hsize=0.7\hsize}X|}}
	\hline
	No.	& & Humanlikeness & Eeriness & Attractiveness \\
		\hline
	 H1 & unbeseelt – lebendig & 0.779 & & \\
	 H2 & synthetisch – echt & 0.845 & & \\
	 H3 & mechanische Bewegungen –  \newline biologische Bewegungen & 0.862 & &\\
	 H4 & von Menschen gemacht – wie ein Mensch & 0.851 & & \\
	 H5 & mit unbegrenzter Lebenszeit – sterblich & 0.494 & &  \\
	 H6 & \textit{künstlich – natürlich} & \textit{0.857} & & \\
	\hline
	E1 & reizlos – irre & & 0.448 & \\
	E2 & vorhersehbar - unheimlich & & 0.642 & \\
	E3 & eindeutig - komisch & & 0.525 &  \\
	E4 & gewöhnlich – ungewöhnlich & & 0.403 & \\
	E5 & langweilig – schockierend & & 0.490 &\\
	E6 & wenig anregend – schaurig & & 0.753 & \\
	E7 & vorhersehbar – spannend & & 0.361 & \\
	E8 & fade – unheimlich & & 0.677 & \\
	E9 & löst keine Emotionen aus – haarsträubend & & 0.646 & \\
	E10 & \textit{beruhigend – unheimlich} & & \textit{0.533} & \\
	\hline
	A1 & hässlich – schön & & & 0.788\\
	A2 & grob – stylisch & & & 0.752\\
	A3 & abstoßend – angenehm & & & 0.711\\
	A4 & unschön – geschmeidig & & & 0.805\\
	A5 & \textit{unattraktiv – attraktiv} & & & \textit{0.812}\\
	\hline
	\end{tabularx}
	\begin{tablenotes}
        \item[\textit{italic}] plausibility question
        \newline Model fit $\chi\textsuperscript{2}=503, df=186,CFI=0.797,SRMR=0.146,RMSEA=0.110$
       	\cite{exploratoryFactorAnalysis}
    \end{tablenotes}	
	\end{threeparttable}
\end{table*}

The eeriness factor had no significant covariance with the other factors (Eeriness-Humanlikeness $p=0.541$; Eeriness-Attractiveness $p=0.443$), but the humanlikeness factor significantly correlated with the attractiveness factor ($p<0.01$), cf.~Table~\ref{tab:cov}. This was already the case in the source translation~\cite{questionnaireHoRevised}. However, since the translation of the attractiveness factor wasn't used for the analysis in our work and therefore wasn't validated by a back translator, a validation and check for this covariance in larger sample sets is still advisable. 

\begin{table}[h!]
    \centering
    \caption{Covariances of factors}\label{tab:cov}
    \begin{tabular}{|c|c|c|}
         \hline
         & Humanlikeness & Attractiveness \\
         \hline
         Eeriness & -0.0608 (p=0.541) & -0.0788 (p=0.443) \\
         \hline
         Humanlikeness & & 0.8121 ($p<0.01$) \\
         \hline
    \end{tabular}
\end{table}







\section{Conclusion}
Overall the analysis reported a high reliability of the question items within the translation. The translation of the pair \enquote{without definite lifetime} - \enquote{mortal} may need to be excluded when fragile robots are to be evaluated due to culture specific difference in interpretation.
The attractiveness indices need to be back translated for validation and the covariance with humanlikeness should be checked with a broader range of samples and more participants.
Despite its inconspicuousness in the analysis the word \enquote{weird} may better be translated with \enquote{seltsam}, to avoid double meaning.

With a total of 60 participants the explanatory power of the results is still limited. However, the results show that a German translation of the original questionnaire is possible and can give meaningful results regarding the humanlikeness and eeriness of robots as perceived by German speakers.

\vspace{3cm}



\bibliographystyle{splncs04}
\bibliography{bibliography.bib}
\balance
\end{document}